\documentclass{article}
\usepackage{spconf,amsmath,amssymb,graphicx}
\usepackage{listings}
\usepackage{caption}
\usepackage{subcaption}
\usepackage{hyperref}
\usepackage{balance}
\usepackage{booktabs}
\usepackage{tabularx} 
\usepackage[symbol]{footmisc}
\newcolumntype{Y}{>{\centering\arraybackslash}X}

\title{Through-Wall Imaging based on WiFi Channel State Information}
\name{Julian Strohmayer, Rafael Sterzinger, Christian Stippel, Martin Kampel}
\address{Computer Vision Lab, TU Wien \\ Favoritenstr. 9/193-1, 1040 Vienna, Austria}

\begin{document}
\maketitle

\begin{abstract}
This work presents a seminal approach for synthesizing images from WiFi Channel State Information~(CSI) in through-wall scenarios.
Leveraging the strengths of WiFi, such as cost-effectiveness, illumination invariance, and wall-penetrating capabilities, our approach enables visual monitoring of indoor environments beyond room boundaries and without the need for cameras.
More generally, it improves the interpretability of WiFi CSI by unlocking the option to perform image-based downstream tasks, e.g., visual activity recognition. In order to achieve this crossmodal translation from WiFi CSI to images, we rely on a multimodal Variational Autoencoder~(VAE) adapted to our problem specifics. We extensively evaluate our proposed methodology through an ablation study on architecture configuration and a quantitative/qualitative assessment of reconstructed images. Our results demonstrate the viability of our method and highlight its potential for practical applications.
\end{abstract}


\vspace{0.5mm}
\section{Introduction}
\label{sec:intro}

\begin{figure}[t!]
  \centering
  \begin{subfigure}{0.49\linewidth}
    \includegraphics[width=1.0\linewidth]{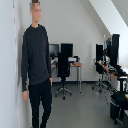}
    \caption{image ground truth}
    \label{teaser:ground_truth}
    \vspace{1mm}
  \end{subfigure}
\hfill
  \begin{subfigure}{0.49\linewidth}
    \includegraphics[width=1.0\linewidth]{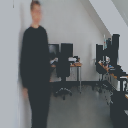}
    \caption{image reconstruction}
    \label{teaser:reconstruction}
    \vspace{1mm}
  \end{subfigure}
  \begin{subfigure}{\linewidth}
    \includegraphics[width=1\linewidth,height=2cm]{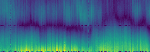}
    \caption{CSI amplitude spectrogram}
    \label{teaser:spectrogram}
  \end{subfigure}
  \vspace{-5mm}
  \caption{Example of an image reconstructed from through-wall WiFi CSI. (a) Shows the ground truth image captured by a camera within the room and (b) the corresponding image reconstruction from (c) the CSI amplitude spectrogram recorded by a WiFi antenna located outside of the room.}
  \vspace{-2mm}
  \label{teaser}
\end{figure}

While Human Behavior Modeling (HBM) has traditionally focused on optical modalities such as RGB, IR, depth, or thermal imaging to meet the requirements of person-centric vision applications~\cite{strohmayer2022compact}, there is a steady rise in the popularity of WiFi as a sensing modality in this domain~\cite{Fu234782379}. Its appeal lies in its capacity for discreet monitoring of vast indoor environments and its practical advantages, such as illumination invariance~\cite{Zhao8578866}, cost-effectiveness~\cite{Schumann2023}, and the protection of visual privacy -- a crucial requirement in privacy-sensitive applications~\cite{Arning2015}.
Despite these advantages, interpreting WiFi signals poses a significant challenge: Deducting human activities or behavior solely based on WiFi Channel State Information (CSI) is inherently difficult, if not impossible (see Figure~\ref{teaser:spectrogram}, which showcases the CSI amplitude spectrogram of a person walking in a room).

To address this problem, our work aims to bridge the gap between WiFi CSI and images.
We propose a new approach that enables visual monitoring in through-wall scenarios and unlocks the potential for image-based downstream tasks based on WiFi CSI, such as the visual annotation of activities.
On top of this, our method also preserves visual privacy as identities are hallucinated based on training data.
Inspired by brain decoding techniques that utilize brain waves for image synthesis~\cite{benchetrit2023brain}, we apply a similar concept to the domain of WiFi CSI. Our approach represents, to our knowledge, the first method for synthesizing images directly from WiFi CSI captured in a through-wall scenario. Leveraging the MoPoE-VAE~\cite{sutter_generalized_2020}, a Variational Autoencoder (VAE) designed for multimodal data, our approach brings the theoretical advantages of WiFi-based person-centric sensing to practical applications. 

The capability of our method to facilitate visual monitoring in through-wall scenarios without conventional cameras is illustrated in Figure~\ref{teaser}.
In this example, the image in Figure~\ref{teaser:reconstruction} is reconstructed solely from the WiFi CSI amplitude spectrogram shown in Figure~\ref{teaser:spectrogram}.
Its close resemblance to the ground truth (Figure~\ref{teaser:ground_truth}) demonstrates the effectiveness of our approach and its potential for enhancing the interpretability of WiFi CSI by unlocking image-based downstream tasks.
To stimulate further research in this emerging field, data, models, and code are made publicly available\footnotemark[1]\footnotetext[1]{Supplementary Material, \href{https://github.com/StrohmayerJ/wificam}{https://github.com/StrohmayerJ/wificam}}.

\section{Related Work}
\label{SecRelatedWork}

Advancements in deep learning and sensing technologies have enabled innovative approaches to extract person-centric information from modalities such as WiFi and radar in through-wall scenarios.
An initial successful attempt that uses WiFi signals in a through-wall setting to estimate human poses is presented by Zhao et al.~\cite{Zhao8578866}, starting a whole line of research that focuses on RF-based HBM.
Subsequently, Li et al.~\cite{Li2020WifiSI} were able to generate binary segmentation masks from WiFi CSI, and  
Hernandez and Bulut~\cite{Hernandez2021} propose an adversarial approach for occupancy monitoring using WiFi CSI.
In continuation of their work, Geng et al.~\cite{geng_densepose_2022} present an advancement in WiFi-based pose estimation, focusing on a multi-person setup, which is, however, limited to intra-room recording.
Another work, by Wu et al.~\cite{Wu2022RFMaskAS}, demonstrates the generation of binary segmentation masks directly from radar signals, adding a new dimension to radar-based synthesis and enabling through-wall object segmentation using radar signals.

Recently, image synthesis from WiFi signals has experienced significant developments.
Notably, the work by Yu et al.~\cite{Yu9949562} demonstrates an innovative two-step process in this domain:
First, human poses are estimated from WiFi CSI and subsequently used to synthesize images, employing a conditional generative adversarial network.
Another approach, although limited to intra-room scenarios, is introduced by Yu et al.~\cite{yu2022rfgan}.
Here, they present a direct method for synthesizing images from radar signals, marking the first attempt in this specific field.
It is important to note that neither of these approaches was applied in through-wall scenarios, leaving a gap that is closed in this work: direct WiFi CSI-based image synthesis in through-wall scenarios.

Benchetrit et al.~\cite{benchetrit2023brain} present an approach for synthesizing images from electroencephalography (EEG) measurements extracted from a human brain.
In their approach, they map images and EEG spectrograms into a shared latent space and hallucinate images using a latent diffusion model.
VAEs have emerged as a powerful tool in multimodal unsupervised learning, which has experienced drastic improvements over the years.
For instance, Wu et al.~\cite{wu_multimodal_2018} enhance the efficacy of VAEs by assuming the multimodal joint posterior is a product of unimodal posteriors.
In this way, information can be aggregated from any subset of unimodal posteriors and can, therefore, be used efficiently when dealing with missing modalities.
However, their technique no longer guarantees a valid lower bound on the joint log-likelihood.
Expanding on this, Shi et al.~\cite{shi_variational_2019} assume that the joint posterior is a mixture of unimodal posteriors, facilitating translation between pairs of modalities.
However, it lacks the capability to leverage the presence of multiple modalities, as it solely considers unimodal posteriors during training.
Finally, Sutter et al.~\cite{sutter_generalized_2020} overcome the limitations of the previous two methodologies, offering efficient many-to-many translation while preserving a valid lower bound on the joint log-likelihood.
Our work is grounded in these foundational studies, leveraging the advancements in VAEs for multimodal unsupervised learning to bridge WiFi CSI signals with images.

\begin{figure}[t!]
  \centering
  \includegraphics[width=\linewidth]{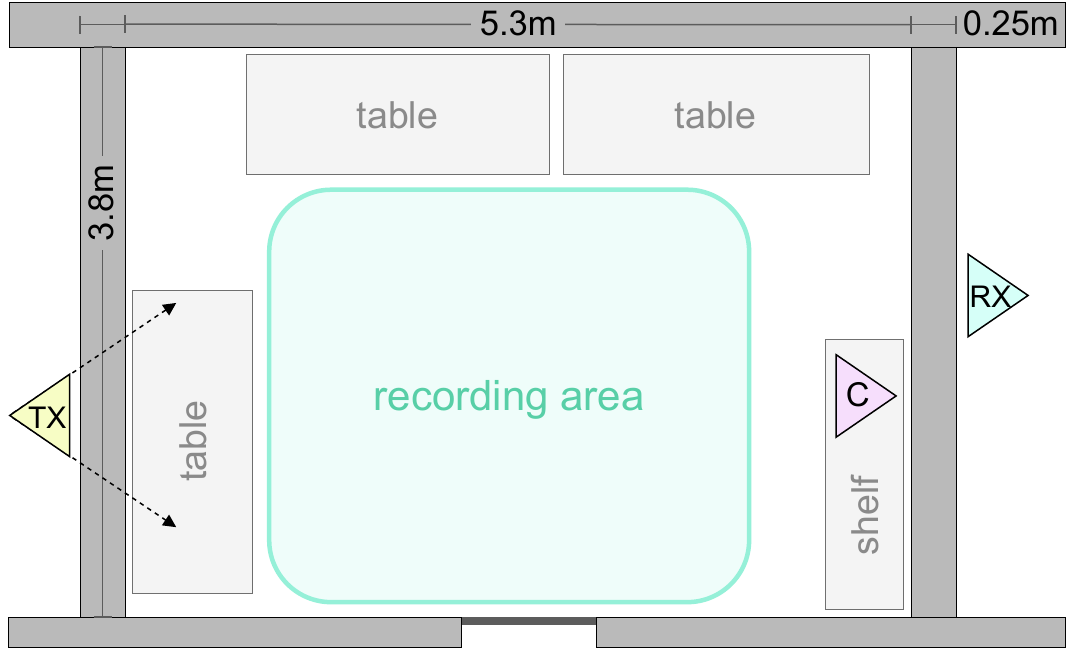}
  \vspace{-4mm}
  \caption{Overview of the experimental setup, showing the arrangement of WiFi transmitter (TX), WiFi receiver (RX), and RGB camera (C) in the office recording environment.}
  \vspace{-2mm}
  \label{environment}
\end{figure}

\section{Data}
\label{SecData}
Our experimental setup, illustrated in Figure~\ref{environment}, is designed to jointly capture through-wall WiFi CSI and images of activities within a target room (3.8m$\times$5.3m office).
In this setup, a point-to-point WiFi transmitter-receiver arrangement is positioned outside the room, accompanied by an intra-room camera.
The WiFi transmitter and receiver systems utilize the \setcounter{footnote}{1}\textit{Espressif ESP32-S3-DevKitC-1U}\footnote{ESP32-S3-DevKitC-1U, \href{https://docs.espressif.com/projects/esp-idf/en/latest/esp32s3/hw-reference/esp32s3/user-guide-devkitc-1.html}{https://docs.espressif.com/}, (acc. 21-01-2024)} development board, equipped with an \textit{ALFA Network APA-M25}\footnote{APA-M25, \href{https://alfa-network.eu/apa-m25}{https://alfa-network.eu/apa-m25}, (acc. 21-01-2024)} directional panel antenna featuring a 66$^{\circ}$ horizontal beam width and a gain of 8dBi @2.4GHz.
The connection between transmitter and receiver is established using the \textit{Espressif ESP-NOW}\footnote{ESP-NOW, \href{https://www.espressif.com/en/solutions/low-power-solutions/esp-now}{https://docs.espressif.com} (acc. 21-01-2024)} protocol, employing a fixed packet-sending rate of 100Hz.
Images are captured at a resolution of 640$\times$480 pixels and a frame rate of 30Hz using an \textit{ESP32-S3}-based camera board\footnote{ESP32-S3-CAM, \href{https://github.com/Freenove/Freenove\_ESP32\_S3\_WROOM\_Board}{https://github.com/} (acc. 21-01-2024)}.
To ensure temporal synchronization, both the WiFi receiver and camera are connected to a notebook that concurrently captures WiFi packets and images through serial over USB and WiFi connections, respectively.

\begin{figure}[t!]
  \centering
  \begin{subfigure}{0.725\linewidth}
    \includegraphics[width=1.0\linewidth]{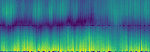}
    \caption{CSI amplitude spectrogram}
    \vspace{-1mm}
  \end{subfigure}
\hspace{-1mm}
  \begin{subfigure}{0.2525\linewidth}
    \includegraphics[width=1.0\linewidth]{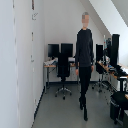}
    \caption{image}
    \vspace{-1mm}
  \end{subfigure}
  \caption{Example of a CSI amplitude spectrogram showing the amplitudes of 52 L-LTF subcarriers over a $\sim$1.5-second time interval and the image corresponding to the central packet.}
  \vspace{-2mm}
  \label{example}
\end{figure}

For training and evaluation purposes, we create a dataset of temporally synchronized CSI and image pairs by recording a ten-minute sequence of continuous walking activities in the test environment, utilizing the described hardware setup.
The raw time series resulting from this comprises 57,413 WiFi packets and 18,261 images, i.e., around three WiFi packets received per image.
Using a visual cue, we eliminate excessive samples at the start and end of the sequence that do not depict the target activity as part of cleaning the raw time series.
Additionally, to account for the sampling rate difference, each WiFi packet is paired with the image having the closest timestamp.
Finally, the dataset is partitioned into training, validation, and test subsets using an 8:1:1 split ratio.
Figure~\ref{example} provides an example of a CSI amplitude spectrogram showing the amplitudes of 52 Legacy Long Training Field (L-LTF) subcarriers over a $\sim$1.5-second time interval (150 WiFi packets) alongside the image corresponding to the central packet.
Our data is made publicly available to ease further research\footnotemark[1].

\vspace{2mm}
\section{Architecture}
\label{SecArchitecture}

In order to infer images from WiFi CSI, we employ MoPoE-VAE~\cite{sutter_generalized_2020}, a multimodal VAE to learn a posterior distribution over a joint latent variable $\mathbf{z} \in \mathbb{R}^D$ given our two modalities, as well as a corresponding decoder to reconstruct images from a sampled latent vector.

Given a set of $\mathbb{X}:={\big\{ \mathbb{X}^{i}\big\}}_{i=1}^N$ samples of images at hand, each paired with a fixed time interval of WiFi packets (the central packet determines the corresponding image), i.e., $\mathbb{X}^{i} = \big\{\mathbf{X}^i_{I},~\mathbf{X}^i_{W}\big\}$, we aim to maximize the log-likelihood of our data at hand:
$$\log p_\theta(\mathbb{X})=\log p_\theta \big(\big\{\mathbb{X}^i\big\}^N_{i=1}\big)$$
Within VAEs, this is achieved by maximizing a lower bound to this objective, the so-called evidence lower bound.
For a given sample $\mathbb{X}^i$, this lower bound on the marginal log-likelihood takes on the following form:
\begin{equation*}
\begin{split}
\mathcal{L}(\theta,\phi; \mathbb{X}^i) := \mathbb{E}_{q_\phi(\mathbf{z}|\mathbb{X}^i)}\big[\log\big( p_\theta(\mathbb{X}^i|\mathbf{z})\big)\big]\\ 
- \beta D_{KL}\big(q_\phi(\mathbf{z}|\mathbb{X}^i)||p_\theta(\mathbf{z})\big),
\end{split}
\end{equation*}
where $D_{KL}$ denotes the Kullback-Leibler divergence ~\cite{kullback1951information} between the approximated posterior and the assumed, in our case, Gaussian prior, and $\beta$ being an additional weight parameter (see Higgins et al. ~\cite{higgins_-vae_2017}), which promotes disentanglement of the latent variable $\mathbf{z}$, in the case that $\beta > 1$. During inference, when image data is missing, i.e.~$\mathbb{X}^i_W:=\mathbb{X}^i\setminus\big\{\mathbf{X}^i_I\big\}$, we still aim to obtain a valid lower bound on the joint probability $\log p_\theta(\mathbb{X}^i)$.
However, when using $\mathcal{L}(\theta, \phi;\mathbb{X}^i_W)$, it only yields a lower bound on $\log p_\theta(\mathbb{X}^i_W)$.
Hence, to obtain a correct lower bound on the joint probability, the following adapted lower bound is used:
\begin{equation*}
\begin{split}
\mathcal{L}_W(\theta,\phi_W; \mathbb{X}^i) := \mathbb{E}_{\tilde{q}_{\phi_W}(\mathbf{z}|\mathbb{X}_W^i)}\big[\log \big(p_\theta(\mathbb{X}^i|\mathbf{z})\big)\big] \\ 
- \beta D_{KL}\big(\tilde{q}_{\phi_W}(\mathbf{z}|\mathbb{X}_W^i)||p_\theta(\mathbf{z})\big).
\end{split}
\end{equation*}

In the general naive case with $M$ different modalities, approximating a lower bound of the joint probability in any case of missing modalities requires the optimization of $2^M$ different models, one for each subset contained within the powerset, posing a drastic scalability issue.
Compared to prior literature, Sutter et al.~\cite{sutter_generalized_2020} circumvent this by modeling the joint posterior approximation as a so-called Mixture of Products of Experts (MoPoE), combining Product of Experts (PoE)~\cite{wu_multimodal_2018} and Mixture of Experts (MoE)~\cite{shi_variational_2019} through abstract mean functions~\cite{e21050485}, to enable efficient retrieval of the joint posterior for all subsets. For more information, we refer the reader to consult~\cite{sutter_generalized_2020}.

While our task at hand considers only two modalities (WiFi CSI, images) and we focus solely on the reconstruction of images from WiFi CSI, we still opt for the same methodology to allow for additional modalities in the future. However, we employ custom VAEs designed for inference from WiFi CSI and adapt the loss to neglect the possibility of decoding a sequence of images, as it is not required.

\begin{figure*}[t!]
  \centering
  \includegraphics[width=\linewidth]{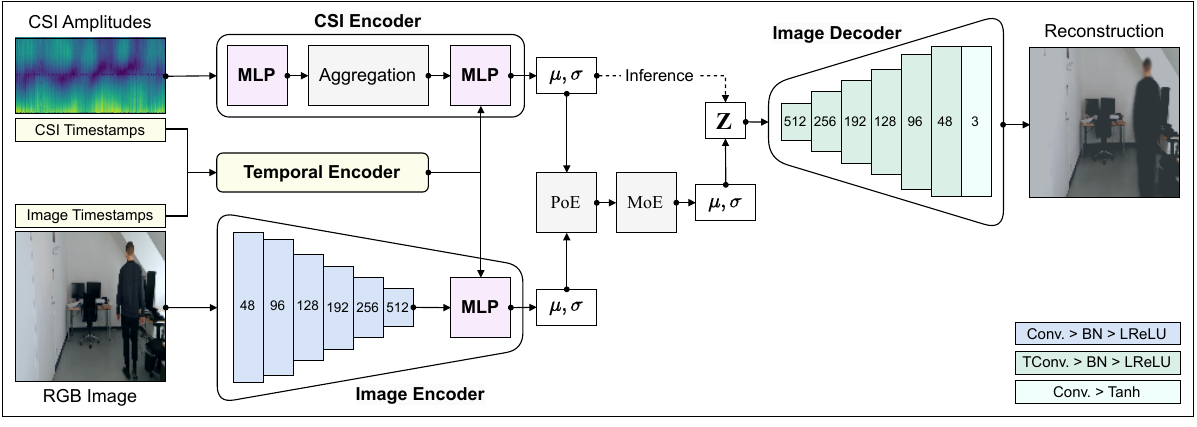}
  \vspace{-5mm}
  \caption{Proposed MoPoE-VAE architecture for WiFi CSI-based image synthesis.}
  \label{pipeline}
  \vspace{-2mm}
\end{figure*}

\subsection{Image Reconstruction}
Given the stark difference of our modalities, we require two architectural different models for learning the unimodal posterior distributions $q_{\phi_I}(\mathbf{z}|\mathbb{X}_I^i)$ and $q_{\phi_W}(\mathbf{z}|\mathbb{X}_W^i)$:

\textbf{Image VAE.} For the encoding and decoding of images, we employ convolutional and transpose-convolutional layers, respectively.
Additionally, we follow each of these layers with batch normalization and a Leaky ReLU activation.
Before inferring the distribution parameters of the latent variable $\mathbf{z}$ given images using the encoder, we rescale the input from a resolution of $640\times480$ down to $128\times128$~pixels in order to reduce computational complexity and apply normalization using per-channels means and standard deviations.
Next, six consecutive convolutional blocks increase the channel size from three (RGB) to 512, followed by a simple Multi-Layer Perceptron (MLP) to map from the flattened output of the convolutions to the unimodal distribution parameters.
Decoding is performed, taking a latent vector $\mathbf{z}$ and reversing the process using transpose-convolutions.

\textbf{CSI VAE.} Concerning our CSI VAE, we first extract information about the signal's amplitude from the raw WiFi CSI and apply an MLP to embed the input per sample of a given sequence to obtain a richer feature representation.
Next, we explore the use of different/no aggregation options along the time dimension before estimating the distribution parameters: uniform feature weighing, Gaussian feature weighing, and concatenation, i.e., flattening the output and leaving it unmodified.
We describe these options in detail below; their effectiveness is evaluated within an ablation study (see Section~\ref{SecEvaluation}).
Lastly, we take the aggregated/concatenated output and employ another MLP to obtain the unimodal distribution parameters of WiFi CSI.
Finally, note that we neglect the reconstruction of WiFi CSI from a latent vector $\mathbf{z}$ as it is not relevant to our task.

During training, we combine the unimodal distribution parameters of our two VAEs, first, to subsets using PoE, and second, combine all subsets within the powerset using MoE.
With predicted parameters $\mathbf{\mu}, \mathbf{\sigma} \in \mathbb{R}^D$, we then sample a latent vector $\mathbf{z}$, using the reparametrization trick; $\mathbf{z}=\mu+\sigma\odot\epsilon$, with $\epsilon\sim\mathcal{N}(0,1)$.
At inference, we only consider the mode of the approximated posterior distribution, predicted by the CSI VAE, for reconstruction, i.e., the predicted mean.

An overview of our proposed methodology for synthesizing images from through-wall WiFi CSI is provided in Figure~\ref{pipeline}, illustrating both training and inference.

\subsection{Aggregation Options}
\vspace{1mm}
As noted previously, regarding the CSI VAE, we consider different aggregations to find the most meaningful representation for predicting an image (determined by the central WiFi packet) from a given sequence.

Let $\tilde{\mathbf{X}}_W \in \mathbb{R}^{L \times H}$ be the embedded WiFi CSI amplitudes after applying the first encoder MLP, where $L$ is the sequence length, and $H$ is the hidden dimension.
First, as a baseline, we consider a \textit{uniform weighing} of our features, i.e., giving each packet equal weight and thus neglecting the order of arrival and the importance of the central packet.
Next, to place more importance on the central packet and reduce permutation invariance, we examine a \textit{Gaussian weighing} with $\mu=\sigma=L/2$.
Both of these options are prone to flickering or, in other words, drastic changes between two consecutive frames.
Finally, to overcome this issue, we consider \textit{concatenation}, leaving the features as is to be fully permutation sensitive.
Hence, depending on the employed options, the input to the second encoder MLP lies either in $\mathbb{R}^{LH}$ or $\mathbb{R}^{H}$.

\subsection{Temporal Encoding}
A final adjustment to our architecture is posed by \textit{temporal encoding.}
Although concatenation alleviates the issue of drastic changes between two consecutive frames, a slight jittering artifact remains.
By adding time information using sinusoidal functions, we are able to improve upon this issue, following a similar strategy as employed by NeRF~\cite{mildenhall2021nerf}.
The encoding for a timestamp \( t \) and a set of frequencies \( F \) is defined as:
\begin{align*}
    T(t, F) = \Big[& \sin\Big(\frac{2^0\pi t}{3L}\Big), \cos\Big(\frac{2^0\pi t}{3L}\Big),\ldots,\\
    & \sin\Big(\frac{2^{F-1}\pi t}{3L}\Big), \cos\Big(\frac{2^{F-1}\pi t}{3L}\Big)\Big],
\end{align*}
where \( t \) is scaled by three times the window size $L$ to have contextual time information from before and after the current window.
Specifically, when incorporating \textit{temporal encoding}, we concatenate the encoding with the image/WiFi features stemming from the CNN/MLP.
This concatenated set of features is then fed through the respective MLP to map to the corresponding distribution parameters (see Figure~\ref{pipeline}).

\begin{figure*}[t!]
  \centering
\begin{subfigure}{0.19\linewidth}
    \includegraphics[width=1.0\linewidth]{img/555_GT.png}
    \caption{image ground truth}
    \label{qualitative:gt}
  \end{subfigure}
  \hfill
\begin{subfigure}{0.19\linewidth}
    \includegraphics[width=1.0\linewidth]{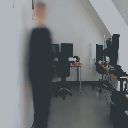}
    \caption{UW}
    \label{qualitative:ue}
  \end{subfigure}
  \hfill
\begin{subfigure}{0.19\linewidth}
    \includegraphics[width=1.0\linewidth]{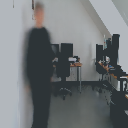}
    \caption{GW}
    \label{qualitative:gw}
  \end{subfigure}
  \hfill
  \begin{subfigure}{0.19\linewidth}
    \includegraphics[width=1.0\linewidth]{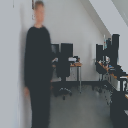}
    \caption{C}
    \label{qualitative:c}
  \end{subfigure}
  \hfill
  \begin{subfigure}{0.19\linewidth}
    \includegraphics[width=1.0\linewidth]{img/555_CT_F.png}
    \caption{C+T}
    \label{qualitative:ct}
  \end{subfigure}
  \caption{Comparison of reconstruction fidelity between MoPoE-VAE models employing the aggregation options (b) \textit{uniform weighing} (UW), (c) \textit{Gaussian weighing} (GW), (d) \textit{concatenation} (C), and (e) \textit{concatenation} with \textit{temporal encoding} (C+T). This visual comparison highlights the improvements in image clarity and reduction of artifacts.}
  \label{qualitativeResults}
  \vspace{-2mm}
\end{figure*}

\section{Evaluation} \label{SecEvaluation}
\vspace{-2mm}
To assess the correlation between reconstruction fidelity and our architectural optimizations, an ablation study is conducted.
In this study, we quantitatively and qualitatively report the reconstruction performance of different model variants, assessing perceived image quality and visual artifacts.

\vspace{-3mm}
\subsection{Model Training}
\vspace{-2mm}

Utilizing the captured dataset, we train models based on the proposed MoPoE-VAE architecture for WiFi-CSI-based image synthesis, considering the following variants: \textit{uniform weighing} (UW), \textit{Gaussian weighing} (GW), and \textit{concatenation} (C), as well as the combination of \textit{concatenation} and \textit{temporal encoding} (C+T).

On top of this, a hyperparameter search is conducted, employing C+T on the validation subset, to determine suitable values for training hyperparameters. We evaluate over batch size $b \in \{16, 32, 64, 128, 256, 512\}$, window size $L \in \{51, 101, 151, 201, 251, 301\}$, and the $D_{KL}$ weighing parameter $\beta \in \{1, 2, 4, 6\}$. This resulted in optimal values of $b=32$, $L=151$, and $\beta=1$, which we subsequently use for training all models.

Concerning the recorded WiFi sequence, for each packet, we extract the 52 L-LTF subcarriers from the complex CSI matrix and precompute amplitudes used to sample $52 \times L$ spectrograms for the CSI VAE to process. Furthermore, the raw images are rescaled to a resolution of 128$\times$128 pixels and normalized using the per-channel means and standard deviations of the dataset. For each configuration, utilizing the Adam optimizer ($lr = 1e-3$), we conduct ten independent training runs spanning 50 epochs and select the model with the lowest validation loss for the final evaluation on the test subset.

\textbf{Metrics.} \label{subMetrics}
To assess the image reconstruction fidelity of our models, we employ metrics well-established in compression/generative model research, such as the Peak Signal-to-Noise Ratio (PSNR), the Structural Similarity Index Measure (SSIM)~\cite{wang2004image}, the Root Mean Squared Error (RMSE), and the Fréchet Inception Distance (FID)~\cite{Heusel2017FID}.

\begin{table}[t!]
\centering
  \caption{Quantitative ablation study results for the aggregation options \textit{uniform weighing} (UW), \textit{Gaussian weighing} (GW), \textit{concatenation} (C), and \textit{concatenation} with \textit{temporal encoding} (C+T). The metrics reported represent the mean and standard deviation across ten independent training runs.}
  \label{tab:quantitativeResults}
  \begin{tabularx}{0.48\textwidth}{p{8mm}>{\centering\arraybackslash}p{14mm}>{\centering\arraybackslash}p{14mm}>{\centering\arraybackslash}p{13mm}>{\centering\arraybackslash}p{15mm}}
    \toprule
    \textbf{Model} &  \textbf{PSNR}$ \uparrow$ & \textbf{SSIM}$\uparrow$ & \textbf{RMSE}$\downarrow$ & \textbf{FID}$ \downarrow$\\
    \midrule
    UW & 20.03$\pm$.05 & 0.734$\pm$.00 & 8.02$\pm$.04 & 142.95$\pm$2.4 \\
    GW & 20.02$\pm$.07 & 0.734$\pm$.00 & 8.00$\pm$.05 & 141.00$\pm$5.0 \\
    C & 20.39$\pm$.05 & 0.748$\pm$.00 & 7.83$\pm$.05 & 127.33$\pm$4.7 \\
    \textbf{C+T} & \textbf{20.39$\pm$.04} & \textbf{0.749$\pm$.00} & \textbf{7.80$\pm$.02} & \textbf{125.62$\pm$3.2} \\
    \bottomrule
  \end{tabularx}
  \vspace{-4mm}
\end{table}

\vspace{-4mm}
\subsection{Quantitative Results}
\vspace{-1mm}
Table~\ref{tab:quantitativeResults} summarizes the quantitative results of our ablation study, providing a comparison between the MoPoE-VAE aggregation options UW, GW, C, and C+T based on the PSNR, SSIM, RMSE, and FID metric, averaged over ten independent training runs.
Among these models, UW and GW are the least effective, yielding comparatively low PSNR, SSIM, and high RMSE and FID scores.

We improve upon this by introducing \textit{concatenation} (C), which turns out to have a significant impact, evidenced by a higher PSNR and SSIM, along with lower RMSE and FID scores.
This stark contrast in quantitative performance is also reflected visually, leading to notable enhancements in sharpness and perceptual fidelity.
However, C+T~(Figure~\ref{qualitative:ct}) outperforms all previous aggregation options, exhibiting slightly higher PSNR and SSIM, coupled with lower RMSE and FID scores, compared to C~(Figure~\ref{qualitative:c}).
Additionally, we point out the observed improvement in FID scores for C+T, suggesting reduced perceptual distance and qualitative enhancements.

Despite the only subtle impact of \textit{temporal encoding} when combined with \textit{concatenation}, reconstructed images are slightly sharper, and, more importantly, frame transitions are more coherent. In summary, while C+T demonstrates superior quantitative performance, its marginal differences highlight the importance of considering both quantitative metrics and visual assessments. 

\subsection{Qualitative Results}
\begin{figure}[t!]
  \centering
\begin{subfigure}{0.19\linewidth}
    \includegraphics[width=1.0\linewidth]{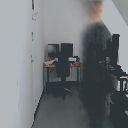}
  \end{subfigure}
  \hfill
\begin{subfigure}{0.19\linewidth}
    \includegraphics[width=1.0\linewidth]{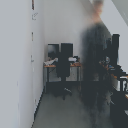}
  \end{subfigure}
  \hfill
\begin{subfigure}{0.19\linewidth}
    \includegraphics[width=1.0\linewidth]{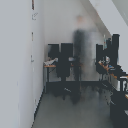}
  \end{subfigure}
  \hfill
  \begin{subfigure}{0.19\linewidth}
    \includegraphics[width=1.0\linewidth]{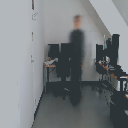}
  \end{subfigure}
  \hfill
  \begin{subfigure}{0.19\linewidth}
    \includegraphics[width=1.0\linewidth]{img/900_GW_d.png}
  \end{subfigure}

\begin{subfigure}{0.19\linewidth}
    \includegraphics[width=1.0\linewidth]{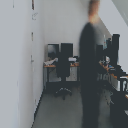}
  \end{subfigure}
  \hfill
\begin{subfigure}{0.19\linewidth}
    \includegraphics[width=1.0\linewidth]{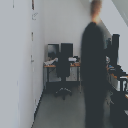}
  \end{subfigure}
  \hfill
\begin{subfigure}{0.19\linewidth}
    \includegraphics[width=1.0\linewidth]{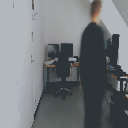}
  \end{subfigure}
  \hfill
  \begin{subfigure}{0.19\linewidth}
    \includegraphics[width=1.0\linewidth]{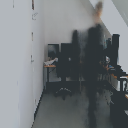}
  \end{subfigure}
  \hfill
  \begin{subfigure}{0.19\linewidth}
    \includegraphics[width=1.0\linewidth]{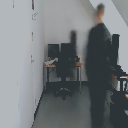}
  \end{subfigure}

  \begin{subfigure}{0.19\linewidth}
    \includegraphics[width=1.0\linewidth]{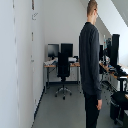}
    \vspace{-6mm}
    \caption*{$i_{1}$}
  \end{subfigure}
  \hfill
\begin{subfigure}{0.19\linewidth}
    \includegraphics[width=1.0\linewidth]{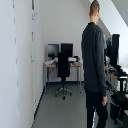}
    \vspace{-6mm}
    \caption*{$i_{25}$}
  \end{subfigure}
  \hfill
\begin{subfigure}{0.19\linewidth}
    \includegraphics[width=1.0\linewidth]{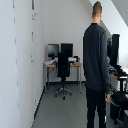}
    \vspace{-6mm}
    \caption*{$i_{50}$}
  \end{subfigure}
  \hfill
  \begin{subfigure}{0.19\linewidth}
    \includegraphics[width=1.0\linewidth]{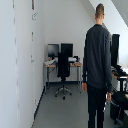}
    \vspace{-6mm}
    \caption*{$i_{75}$}
  \end{subfigure}
  \hfill
  \begin{subfigure}{0.19\linewidth}
    \includegraphics[width=1.0\linewidth]{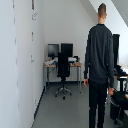}
    \vspace{-6mm}
    \caption*{$i_{100}$}
  \end{subfigure}

  \begin{subfigure}{\linewidth}
    \includegraphics[width=1.0\linewidth]{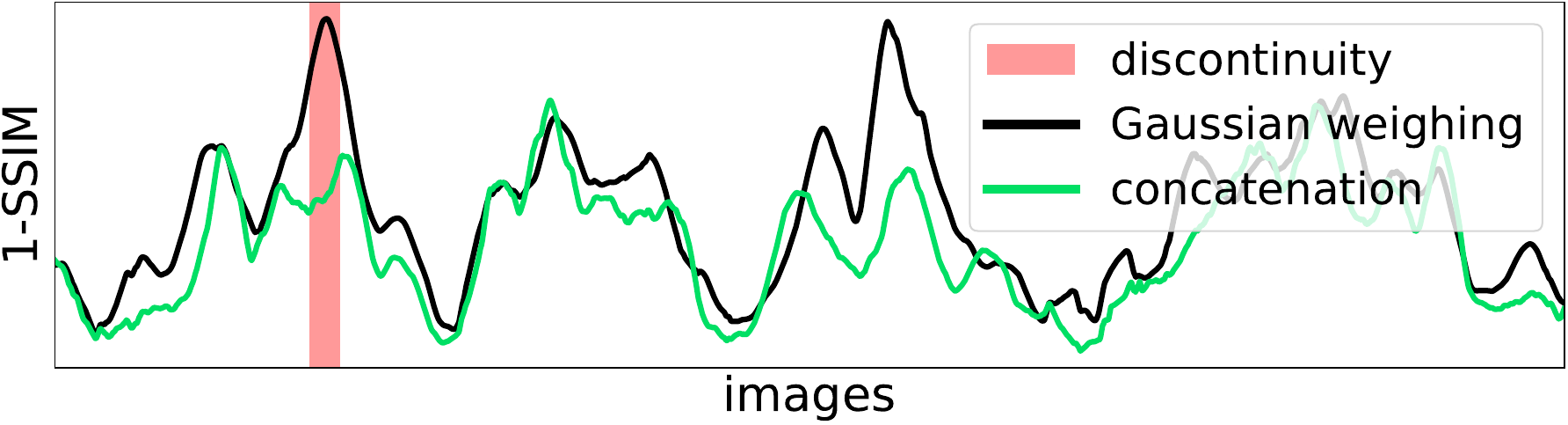}
    \vspace{-6mm}
  \end{subfigure}
  \vspace{-5mm}
  \caption{Example showing the elimination of spatiotemporal discontinuities in a sequence of 100 test images with high 1-SSIM, highlighted in red, through the aggregation option \textit{concatenation}. From top to bottom, the rows show \textit{Gaussian weighing} (GW), \textit{concatenation} (C), and ground truth, respectively.}
  \vspace{-5mm}
  \label{jumpResults}
\end{figure}

\vspace{-1mm}
Our qualitative examination aligns with the trends observed in the quantitative analysis;
Figure~\ref{qualitativeResults} illustrates the impact of different aggregation options on image reconstruction fidelity.
In Figure~\ref{qualitative:ue}, the \textit{uniform weighing} (UW) model yields a blurry reconstruction, consistent with its lower PSNR and higher RMSE scores (see Table~\ref{tab:quantitativeResults}).
We hypothesize that the lack of focus in the temporal domain results in diminished perceptual quality.
Moving to Figure~\ref{qualitative:gw}, the \textit{Gaussian weighting} (GW) model showcases a slight improvement in quality despite the similarity in quantitative metrics.
This improvement is attributed to the prioritization of central WiFi packets during weighing; that is, packets that are in closer temporal proximity to the ground truth image have a higher contribution, resulting in a clearer and more accurate reconstruction.
Next, Figure~\ref{qualitative:c} illustrates the \textit{concatenation} (C) model: By learning the aggregation of WiFi features itself, not only further improves the sharpness of images but even enhances the quality on a frame-to-frame level, eliminating spatiotemporal discontinuities, as illustrated in Figure~\ref{jumpResults}.
Quantitatively, C outperforms UW and GW, yet subtle artifacts and jittering remain visible in videos compared to C+T, possibly contributed by marginal differences in FID scores.
Finally, the peak of our architectural improvements is reached when employing the C+T model, which is showcased in Figure~\ref{qualitative:ct}, incorporating both \textit{concatenation} and \textit{temporal encoding}.
Employing cyclic \textit{temporal encoding} effectively mitigates the previously mentioned jittering artifact, resulting in smoother frame transitions and reduced FID scores.
For a perspective on video quality, reconstructed videos complement these static images, showcasing for each aggregation option its ability to reconstruct images based on through-wall WiFi CSI\footnotemark[1].

In conclusion, these qualitative results demonstrate the capability of our method to facilitate through-wall visual monitoring without conventional cameras. The successful reconstruction of images from WiFi CSI not only validates our approach but also suggests its potential for image-based downstream tasks. This represents a significant step toward enhancing the interpretability of WiFi CSI, providing a foundation for further exploration and research in this evolving field.

\vspace{-3mm}
\section{Limitations} \label{SecFutureWork}
\vspace{-2mm}
While the MoPoE-VAE+C+T model shows promising results on our dataset, demonstrating the feasibility of image synthesis from through-wall WiFi CSI, it may not generalize to new people or environments without further adaptations. Generalization is an open problem in WiFi-based person-centric sensing, requiring additional steps across diverse scenarios~\cite{chen2023cross}. Enhancing cross-domain generalization could involve domain-invariant feature extraction techniques or training data from diverse domains without needing architectural changes. Additionally, capturing samples in the target domain for fine-tuning or test-time training could improve generalization in practical applications. Another promising direction for future work is expanding the modalities used in our approach. The MoPoE-VAE framework is adaptable to multimodal data, allowing for additional sensory inputs. Combining WiFi CSI with radar, audio, or structural vibrations could provide complementary information and help in scenarios where one modality alone is insufficient for accurate reconstructions, such as when the WiFi signal is disturbed or ambiguous due to environmental effects.

\vspace{-3mm}
\section{Conclusion} \label{SecConclusion}
\vspace{-3mm}
In this work, we have demonstrated the direct synthesis of person-centric images from WiFi CSI captured in a through-wall scenario. We collected a dataset of WiFi CSI and corresponding image time series, based on which an ablation study was conducted to assess the fidelity of reconstructed images stemming from different architectural variations. From this evaluation, a model pairing with the aggregation option \textit{concatenation} and \textit{temporal encoding} emerged as superior. Our results demonstrate the viability of the proposed approach, suggesting its potential for novel applications, such as through-wall visual monitoring without conventional cameras, and improving the interpretability of WiFi CSI by unlocking image-based downstream tasks, such as visual activity recognition.

\vspace{-4mm}
\section{Acknowledgments} \label{SecAcknowledment}
\vspace{-3mm}
This work is partly funded by the Vienna Business Agency (grant 4829418) and the Austrian security research program KIRAS of the Austrian Research Promotion Agency FFG (grant 49450173).

\bibliographystyle{IEEEbib}
\bibliography{refs}

\begin{thebibliography}{10}

\bibitem{strohmayer2022compact}
Julian Strohmayer and Martin Kampel,
\newblock ``A compact tri-modal camera unit for rgbdt vision,''
\newblock in {\em 2022 the 5th International Conference on Machine Vision and Applications (ICMVA)}, 2022, pp. 34--42.

\bibitem{Fu234782379}
Biying Fu, Naser Damer, Florian Kirchbuchner, and Arjan Kuijper,
\newblock ``Sensing technology for human activity recognition: A comprehensive survey,''
\newblock {\em IEEE Access}, vol. PP, pp. 1--1, 01 2020.

\bibitem{Zhao8578866}
Mingmin Zhao, Tianhong Li, Mohammad~Abu Alsheikh, Yonglong Tian, Hang Zhao, Antonio Torralba, and Dina Katabi,
\newblock ``Through-wall human pose estimation using radio signals,''
\newblock in {\em 2018 IEEE/CVF Conference on Computer Vision and Pattern Recognition}, 2018, pp. 7356--7365.

\bibitem{Schumann2023}
Robert Schumann, Frédéric Li, and Marcin Grzegorzek,
\newblock ``Wifi sensing with single-antenna devices for ambient assisted living,''
\newblock 10 2023, pp. 1--8.

\bibitem{Arning2015}
Katrin Arning and Martina Ziefle,
\newblock ````get that camera out of my house!'' conjoint measurement of preferences for video-based healthcare monitoring systems in private and public places,''
\newblock in {\em Inclusive Smart Cities and e-Health}, Antoine Geissb{\"u}hler, Jacques Demongeot, Mounir Mokhtari, Bessam Abdulrazak, and Hamdi Aloulou, Eds., Cham, 2015, pp. 152--164, Springer International Publishing.

\bibitem{benchetrit2023brain}
Yohann Benchetrit, Hubert Banville, and Jean-Rémi King,
\newblock ``Brain decoding: toward real-time reconstruction of visual perception,'' 2023.

\bibitem{sutter_generalized_2020}
Thomas~M. Sutter, Imant Daunhawer, and Julia~E. Vogt,
\newblock ``Generalized multimodal {ELBO},''
\newblock in {\em International Conference on Learning Representations}, 2020.

\bibitem{Li2020WifiSI}
Chenning Li, Zheng Liu, Yuguang Yao, Zhichao Cao, Mi~Zhang, and Yunhao Liu,
\newblock ``Wi-fi see it all: generative adversarial network-augmented versatile wi-fi imaging,''
\newblock {\em Proceedings of the 18th Conference on Embedded Networked Sensor Systems}, 2020.

\bibitem{Hernandez2021}
Steven~M. Hernandez and Eyuphan Bulut,
\newblock ``Adversarial occupancy monitoring using one-sided through-wall wifi sensing,''
\newblock in {\em ICC 2021 - IEEE International Conference on Communications}, 2021, pp. 1--6.

\bibitem{geng_densepose_2022}
Jiaqi Geng, Dong Huang, and Fernando De~la Torre,
\newblock ``{DensePose} from {WiFi},'' .

\bibitem{Wu2022RFMaskAS}
Zhi Wu, Dongheng Zhang, Chunyang Xie, Cong Yu, Jinbo Chen, Yang Hu, and Yan Chen,
\newblock ``Rfmask: A simple baseline for human silhouette segmentation with radio signals,''
\newblock {\em IEEE Transactions on Multimedia}, vol. 25, pp. 4730--4741, 2022.

\bibitem{Yu9949562}
Cong Yu, Dongheng Zhang, Chunyang Xie, Zhi Lu, Yang Hu, Houqiang Li, Qibin Sun, and Yan Chen,
\newblock ``Wifi-based human pose image generation,''
\newblock in {\em 2022 IEEE 24th International Workshop on Multimedia Signal Processing (MMSP)}, 2022, pp. 1--6.

\bibitem{yu2022rfgan}
Cong Yu, Zhi Wu, Dongheng Zhang, Zhi Lu, Yang Hu, and Yan Chen,
\newblock ``Rfgan: Rf-based human synthesis,''
\newblock {\em IEEE Transactions on Multimedia}, 2022.

\bibitem{wu_multimodal_2018}
Mike Wu and Noah Goodman,
\newblock ``Multimodal generative models for scalable weakly-supervised learning,''
\newblock in {\em Advances in Neural Information Processing Systems}. 2018, vol.~31, Curran Associates, Inc.

\bibitem{shi_variational_2019}
Yuge Shi, Siddharth N, Brooks Paige, and Philip Torr,
\newblock ``Variational mixture-of-experts autoencoders for multi-modal deep generative models,''
\newblock in {\em Advances in Neural Information Processing Systems}. 2019, vol.~32, Curran Associates, Inc.

\bibitem{kullback1951information}
Solomon Kullback and Richard~A Leibler,
\newblock ``On information and sufficiency,''
\newblock {\em The annals of mathematical statistics}, vol. 22, no. 1, pp. 79--86, 1951.

\bibitem{higgins_-vae_2017}
Irina Higgins, Loic Matthey, Arka Pal, Christopher Burgess, Xavier Glorot, Matthew Botvinick, Shakir Mohamed, and Alexander Lerchner,
\newblock ``$\beta$-{VAE}: {Learning} {Basic} {Visual} {Concepts} {with} a {Constrained} {Variational} {Framework},''
\newblock 2017.

\bibitem{e21050485}
Frank Nielsen,
\newblock ``On the jensen–shannon symmetrization of distances relying on abstract means,''
\newblock {\em Entropy}, vol. 21, no. 5, 2019.

\bibitem{mildenhall2021nerf}
Ben Mildenhall, Pratul~P Srinivasan, Matthew Tancik, Jonathan~T Barron, Ravi Ramamoorthi, and Ren Ng,
\newblock ``Nerf: Representing scenes as neural radiance fields for view synthesis,''
\newblock {\em Communications of the ACM}, vol. 65, no. 1, pp. 99--106, 2021.

\bibitem{wang2004image}
Zhou Wang, Alan~C Bovik, Hamid~R Sheikh, and Eero~P Simoncelli,
\newblock ``Image quality assessment: from error visibility to structural similarity,''
\newblock {\em IEEE transactions on image processing}, vol. 13, no. 4, pp. 600--612, 2004.

\bibitem{Heusel2017FID}
Martin Heusel, Hubert Ramsauer, Thomas Unterthiner, Bernhard Nessler, and Sepp Hochreiter,
\newblock ``Gans trained by a two time-scale update rule converge to a local nash equilibrium,'' 2018.

\bibitem{chen2023cross}
Chen Chen, Gang Zhou, and Youfang Lin,
\newblock ``Cross-domain wifi sensing with channel state information: A survey,''
\newblock {\em ACM Computing Surveys}, vol. 55, no. 11, pp. 1--37, 2023.

\end{thebibliography}

\end{document}